\documentclass[accepted]{uai2026} 

\usepackage[american]{babel}
\usepackage{float}
\usepackage{placeins} 
\usepackage{natbib} 
\usepackage{mathtools} 
\usepackage{booktabs} 
\usepackage{graphicx} 
\usepackage{tikz} 
\usepackage{amssymb}

\title{Neural Conjugate Aggregation: Identifiable Unsupervised Multi-Sensor Regression under Heterogeneous Sensor Bias}

\author[1]{Muhammed Faruk Aytin}
\author[1]{Zehra Demir}
\author[2]{Alper Unal}
\author[3]{Julian Marshall}
\author[1]{Gozde Unal}
\affil[1]{%
    AI and Data Engineering Department\\
    Istanbul Technical University\\
    Istanbul, Türkiye
}
\affil[2]{%
    Eurasia Institute of Earth Sciences\\
    Istanbul Technical University\\
    Istanbul, Türkiye
}
\affil[3]{%
    Civil \& Environmental Engineering\\
    University of Washington\\
    Seattle, WA, USA
  }

\begin{document}
\maketitle

\begin{abstract}
We study regression-based data fusion under uncertainty, where multiple noisy and biased measurement sources are available but ground-truth labels are absent during training. This setting arises in sensor networks, simulation ensembles, and scientific monitoring systems where supervision is costly or infeasible.
We propose the Neural Conjugate Aggregation Model (NCAM), a hierarchical Bayesian framework that combines neural networks with conjugate Gaussian inference for unsupervised multi-source fusion. NCAM learns source-specific bias and reliability conditioned on contextual covariates, yielding an analytically tractable posterior over a latent target variable with decomposed epistemic and aleatoric uncertainty. Structural non-identifiability is resolved through sensor anchoring and variance regularization, enabling stable and interpretable posterior aggregation.
To complement Bayesian uncertainty with finite-sample guarantees, we integrate locally adaptive Monte Carlo conformal prediction, producing heteroscedastic prediction intervals with coverage guarantees under exchangeability assumptions. Experiments on synthetic and real-world air-quality datasets demonstrate improved predictive accuracy and well-calibrated uncertainty compared to unsupervised baselines, including mean aggregation, probabilistic PCA, and Kalman filtering.

\end{abstract}

\section{Introduction}\label{sec:intro}

Multi-sensor systems are increasingly used to monitor complex physical, environmental, and engineered processes. Examples include air-quality monitoring networks, industrial sensing systems, autonomous platforms, and scientific observation infrastructures. In these settings, multiple sensors provide complementary measurements of an underlying latent quantity of interest. Combining these observations into a coherent estimate is the central goal of sensor fusion.

Classical sensor fusion methods typically assume that sensor characteristics are known or can be calibrated using labeled data. In practice, however, these assumptions are often violated. Low-cost sensors may exhibit unknown biases, heterogeneous noise levels, environmental confounding, temporal drift, and varying reliability across operating conditions. At the same time, obtaining ground-truth labels from high-quality reference instruments is frequently expensive, sparse, or entirely infeasible. As a result, many real-world sensing systems operate in a regime where multiple noisy observations are available, but direct supervision is not.

This motivates the problem of \emph{unsupervised multi-sensor regression}: inferring an underlying latent target from heterogeneous sensor measurements without access to target labels during training. While seemingly simple, this setting introduces several fundamental challenges.

First, sensor-specific biases and reliabilities must be learned solely from the relationships among sensors themselves, rather than from comparisons against known targets. Second, latent-variable fusion models suffer from structural non-identifiability: multiple parameterizations may explain the same observations equally well, making the recovered latent state ambiguous. Third, uncertainty quantification becomes particularly challenging because neither model calibration nor uncertainty calibration can rely on labeled target observations. Finally, these challenges must be addressed simultaneously rather than in isolation, since sensor bias, identifiability, and uncertainty estimation are tightly coupled.

Existing approaches typically address only subsets of these difficulties. Supervised calibration methods rely on access to reference measurements. Classical Bayesian fusion methods often assume known sensor models or calibrated measurement processes. Heuristic aggregation strategies, such as averaging or inverse-variance weighting, do not explicitly resolve latent-state identifiability. More recently, uncertainty quantification methods such as conformal prediction provide finite-sample coverage guarantees, but generally assume access to labeled calibration targets. To our knowledge, there is currently no unified framework that simultaneously addresses unsupervised sensor calibration, heterogeneous sensor reliability, latent-state identifiability, and uncertainty quantification in the absence of target labels.

In this work, we propose the \emph{Neural Conjugate Aggregation Model (NCAM)}, a probabilistic framework for unsupervised multi-sensor regression. NCAM combines context-dependent neural sensor models with a hierarchical Bayesian latent-variable formulation that admits closed-form conjugate posterior aggregation. The framework explicitly models sensor-specific bias and reliability, resolves structural non-identifiability through an anchoring mechanism, and produces posterior predictive uncertainty estimates without requiring labeled targets during training.

Importantly, the contribution of NCAM is not the introduction of Bayesian inference, neural networks, or conformal prediction individually. Rather, the contribution lies in a principled formulation that brings these components together to address the unique challenges of unsupervised sensor fusion. This formulation enables sensor bias estimation, reliability-aware aggregation, identifiable latent-state inference, and uncertainty-aware prediction within a single coherent framework.

To complement Bayesian posterior uncertainty, we further introduce a ground-truth-free Monte Carlo conformal calibration procedure that produces locally adaptive prediction intervals using only unlabeled sensor observations. This provides a practical mechanism for uncertainty calibration in settings where traditional supervised conformal methods cannot be directly applied.

Our main contributions are:

\begin{itemize}
\item We formulate unsupervised multi-sensor regression as a probabilistic latent-state inference problem involving heterogeneous sensor biases, reliability estimation, identifiability, and uncertainty quantification.

\item We introduce NCAM, a hierarchical Bayesian framework that learns sensor bias and reliability while admitting closed-form conjugate posterior aggregation.

\item We resolve structural non-identifiability through a principled anchoring strategy, yielding interpretable and identifiable latent-state estimates.

\item We develop a ground-truth-free Monte Carlo conformal calibration procedure for uncertainty quantification without labeled target observations.

\item We demonstrate improved predictive accuracy and uncertainty estimation on both synthetic and real-world air-quality sensing datasets.
\end{itemize}

\section{Related Work}
\label{sec:related}

\paragraph{Sensor fusion and uncertainty-aware aggregation.}
Classical multi-sensor fusion is rooted in probabilistic state estimation, with the Kalman filter providing a canonical Bayesian aggregation rule under linear-Gaussian assumptions \citep{kalman1960filtering}. Modern machine learning approaches instead treat fusion as supervised multi-modal learning, where neural networks learn data-dependent weighting or shared latent representations end-to-end \citep{baltrusaitis2019multimodal}. Recent work further incorporates reliability modeling or evidential formulations to down-weight degraded modalities and improve confidence estimation \citep{cocoon2024robust,huang2025deepevidentialfusion}. However, most of these approaches rely on ground-truth supervision and do not explicitly address structural identifiability or uncertainty calibration in fully unsupervised settings.

\paragraph{Unsupervised fusion under sensor drift.}
In environmental sensing and air-quality monitoring, dense reference labels are often unavailable, and low-cost sensors exhibit time-varying bias due to drift and environmental confounders \citep{shahid}. This has motivated methods that exploit inter-sensor agreement and covariate structure for weakly supervised or unsupervised correction \citep{dalbah2025veli,atmos}. Our work fits this regime but differs in two key aspects: we formulate fusion as a latent-variable generative model with conjugate Bayesian aggregation, and we explicitly address non-identifiability and uncertainty propagation within an unsupervised framework.

\paragraph{Conformal prediction and ambiguous targets.}
Conformal prediction provides finite-sample marginal coverage under exchangeability \citep{angelopoulos}, with locally adaptive extensions for heteroscedastic regression \citep{romano2019cqr} and relaxations under weak dependence \citep{barber2023beyond,enbpi}. Most conformal methods assume access to observed calibration targets. 

Recent work by \citet{stutz2023ambiguous} introduced Monte Carlo conformal prediction under ambiguous ground truth, showing that sampling-based calibration can provide valid prediction sets when labels are uncertain. We build on this insight but consider a fundamentally different setting: continuous regression with a \emph{latent, unobserved target} arising from unsupervised multi-sensor fusion. Unlike \citet{stutz2023ambiguous}, who operate in supervised classification with ambiguous labels, we integrate posterior predictive sampling from a conjugate latent-variable model and address identifiability in the absence of any training labels. This bridges generative Bayesian fusion and ground-truth–free conformal calibration, a combination that has not been explored in prior work.

\section{Method}\label{sec:method}

\subsection{Problem Setup}

We consider unsupervised multi-sensor regression. For each entity $i = 1,\dots,N$, we observe contextual covariates $X_i \in \mathbb{R}^{d_x}$ and $L$ noisy measurements 
$B_i = \{b_{i1}, \dots, b_{iL}\}$ from heterogeneous sensors providing noisy estimates of a common latent target.

The goal is to infer a latent continuous variable $\Theta_i$ representing the fused target value, without access to ground-truth supervision during training. A summary of the key notation used throughout the paper is located in Table~\ref{tab:notation} in Appendix~\ref{app:notation}.


\subsection{Neural Conjugate Aggregation Model (NCAM)}

We model fusion using a hierarchical latent-variable model with directed structure
\[
X_i \;\to\; \Theta_i \;\to\; (B_i, Y_i),
\]
where $\Theta_i$ denotes the latent fused target and $Y_i$ represents the true (unobserved) quantity of interest.

\paragraph{Generative Model.}

\textbf{Latent prior.}
\begin{equation}
p(\Theta_i \mid X_i)
=
\mathcal{N}\!\big(\Theta_i;\, \mu_0(X_i),\, \sigma_0^2(X_i)\big),
\label{eq:prior}
\end{equation}
where $\mu_0(X_i)$ and $\sigma_0^2(X_i)$ are outputs of a neural prior network.

\textbf{Measurement model.}
\begin{equation}
p(B_{ij} \mid \Theta_i, X_i)
=
\mathcal{N}\!\big(
b_{ij};\,
a_j(X_i)\Theta_i + c_j(X_i),\,
\sigma_j^2(X_i)
\big),
\label{eq:likelihood}
\end{equation}
where $a_j(X_i)$ and $c_j(X_i)$ model sensor-specific multiplicative and additive bias, and $\sigma_j^2(X_i)$ captures source-dependent noise.

\textbf{Latent observation model.}
\begin{equation}
p(Y_i \mid \Theta_i)
=
\mathcal{N}\!\big(Y_i;\, \Theta_i,\, \sigma_y^2\big),
\label{eq:truey}
\end{equation}
where $\sigma_y^2$ is a small fixed constant controlling residual aleatoric variability.

Neural networks parameterize the prior mean and variance $(\mu_0, \sigma_0^2)$, the sensor-specific variances $\{\sigma_j^2\}_{j=1}^L$, and the bias terms $\{a_j, c_j\}_{j=1}^L$.

\paragraph{Posterior Aggregation.}

Assuming conditional independence of measurements given $\Theta_i$, the likelihood factorizes as
\[
p(B_i \mid \Theta_i, X_i)
=
\prod_{j=1}^L 
\mathcal{N}\!\left(b_{ij};\, a_j(X_i)\Theta_i + c_j(X_i),\, \sigma_j^2(X_i)\right).
\]

Combining this likelihood with the Gaussian prior yields a conjugate Gaussian posterior:
\begin{equation}
p(\Theta_i \mid B_i, X_i)
=
\mathcal{N}\!\left(\Theta_i;\, \mu_{\text{post}}(B_i,X_i),\, \sigma_{\text{post}}^2(B_i,X_i)\right).
\label{eq:posterior}
\end{equation}

The posterior precision and mean are
\begin{align}
\sigma_{\text{post}}^{-2}(B_i,X_i)
&=
\sigma_0^{-2}(X_i)
+ \sum_{j=1}^L \frac{a_j^2(X_i)}{\sigma_j^2(X_i)},
\label{eq:postprec}\\
\mu_{\text{post}}(B_i,X_i)
&=
\sigma_{\text{post}}^2(B_i,X_i)
\Bigg[
\frac{\mu_0(X_i)}{\sigma_0^2(X_i)} 
\nonumber\\
&\qquad
+ \sum_{j=1}^L
\frac{a_j(X_i)\bigl(b_{ij}-c_j(X_i)\bigr)}{\sigma_j^2(X_i)}
\Bigg].
\label{eq:postmean}
\end{align}

The posterior mean $\mu_{\text{post}}$ provides a reliability-weighted, bias-corrected fusion of all sensors, where each source contributes proportionally to $a_j^2(X_i)/\sigma_j^2(X_i)$.

\paragraph{Identifiability in Multi-sensor Latent Models}
\label{sec:identifiability}

The measurement model
\begin{equation}
b_{ij} = a_j(X_i)\,\Theta_i + c_j(X_i) + \varepsilon_{ij},
\qquad
\varepsilon_{ij} \sim \mathcal{N}(0, \sigma_j^2(X_i)),
\label{eq:meas_model_ident}
\end{equation}
is invariant under affine reparameterizations of the latent variable.
For any $e \neq 0$ and $f$, define
\[
\begin{aligned}
\Theta_i' &= e\,\Theta_i + f, \\
a_j'(X_i) &= \frac{a_j(X_i)}{e}, \\
c_j'(X_i) &= c_j(X_i) - \frac{a_j(X_i)}{e}\,f .
\end{aligned}
\]
so the measurement likelihood is unchanged. Consequently, $(\Theta_i, a_j, c_j)$ are not jointly identifiable.

This structural symmetry allows the marginal likelihood to decrease while the latent scale drifts, potentially degrading predictive RMSE despite improving NLL. Moreover, since conformal calibration operates on predictive distributions, it cannot resolve this latent non-identifiability.

\paragraph{Sensor Anchoring for Identifiable Fusion}
To break the affine symmetry, we fix one reference sensor $j^\star$:
\begin{equation}
a_{j^\star}(X) \equiv 1,
\qquad
c_{j^\star}(X) \equiv 0.
\label{eq:anchor}
\end{equation}
This removes the scale and location degrees of freedom, rendering $\Theta_i$ identifiable relative to the anchor. Posterior precision weights then reflect true reliability differences rather than arbitrary rescalings, ensuring that improvements in marginal likelihood correspond to genuine improvements in predictive accuracy.

\subsection{Posterior Predictive Distribution}

The predictive distribution for the true (unobserved) target variable $Y_i$ is obtained by marginalizing over the latent variable $\Theta_i$:
\begin{equation}
p(Y_i \mid B_i, X_i)
= \int p(Y_i \mid \Theta_i)\, p(\Theta_i \mid B_i, X_i)\, d\Theta_i.
\end{equation}

Substituting the Gaussian forms from Eqs.~\eqref{eq:truey} and \eqref{eq:posterior}, the integral admits a closed-form solution:
\begin{equation}
p(Y_i \mid B_i, X_i)
=
\mathcal{N}\!\left(
Y_i;\,
\mu_{\text{post}}(B_i,X_i),\,
\sigma_{\text{post}}^2(B_i,X_i) + \sigma_y^2
\right).
\label{eq:predictive}
\end{equation}

The predictive mean is therefore
\begin{equation}
\hat{Y}_i^{\text{fused}} = \mu_{\text{post}}(B_i,X_i),
\end{equation}
and the predictive variance decomposes as
\begin{equation}
\mathrm{Var}(Y_i \mid B_i, X_i)
=
\sigma_{\text{post}}^2(B_i,X_i) + \sigma_y^2,
\end{equation}
capturing epistemic uncertainty propagated through the multi-sensor aggregation and aleatoric uncertainty intrinsic to the underlying process.

\subsection{Unsupervised Training via Marginal Likelihood Maximization}

Since the true target values $Y_i^\star$ are not observed, the neural prior and sensor parameters are learned in an unsupervised manner by maximizing the marginal likelihood of the observed multi-source measurements $B_i$ conditioned on covariates $X_i$.

\paragraph{Marginal Likelihood.}
The marginal likelihood is obtained by integrating out the latent variable $\Theta_i$:
\begin{equation}
p(B_i \mid X_i)
=
\int p(B_i \mid \Theta_i, X_i)\, p(\Theta_i \mid X_i)\, d\Theta_i.
\label{eq:marginal}
\end{equation}

From the generative model (Eqs.~\ref{eq:prior} and \ref{eq:likelihood}), the conditional distribution of $B_i$ given $\Theta_i$ is Gaussian:
\begin{multline}
p(B_i \mid \Theta_i, X_i)
= \mathcal{N}_L\!\Big(
B_i;\,
A_i \Theta_i + \mathbf{c}_i,\\
\mathrm{diag}\!\big(\sigma_1^2(X_i),\ldots,\sigma_L^2(X_i)\big)
\Big),
\end{multline}
where
\begin{equation}
B_i =
\begin{bmatrix}
b_{i1}\\
\vdots\\
b_{iL}
\end{bmatrix},
\quad
A_i =
\begin{bmatrix}
a_1(X_i)\\
\vdots\\
a_L(X_i)
\end{bmatrix},
\quad
\mathbf{c}_i =
\begin{bmatrix}
c_1(X_i)\\
\vdots\\
c_L(X_i)
\end{bmatrix}.
\end{equation}

The prior over $\Theta_i$ is given by
\begin{equation}
p(\Theta_i \mid X_i)
=
\mathcal{N}\!\left(\Theta_i;\, \mu_0(X_i),\, \sigma_0^2(X_i)\right).
\end{equation}

Because both terms are Gaussian, the integral in Eq.~\eqref{eq:marginal} admits a closed-form solution:
\begin{equation}
\begin{split}
p(B_i \mid X_i)
&= \mathcal{N}_L\!\Big(
B_i;\,
A_i \mu_0(X_i) + \mathbf{c}_i,\\
&\qquad
\sigma_0^2(X_i)\, A_i A_i^\top
+ \mathrm{diag}\!\big(\sigma_1^2(X_i),\ldots,\sigma_L^2(X_i)\big)
\Big).
\end{split}
\label{eq:marginalGaussian}
\end{equation}

This marginal likelihood captures both shared uncertainty through the rank-one covariance term
$\sigma_0^2(X_i) A_i A_i^\top$ induced by the latent variable $\Theta_i$,
and source-specific noise through the diagonal covariance matrix.

\paragraph{Training Objective.}
The model parameters $\phi$ and $\psi = \{\psi_j\}_{j=1}^L$ are learned by minimizing the negative log marginal likelihood over all entities:
\begin{equation}
\begin{split}
\mathcal{L}_{\mathrm{NCAM}}(\phi,\psi)
=
-\sum_{i=1}^N
\log
\mathcal{N}_L\!\Big(
B_i;\,
A_i \mu_0(X_i) + \mathbf{c}_i,\,
\\
\sigma_0^2(X_i)\, A_i A_i^\top
+
\mathrm{diag}\!\big(\sigma_1^2(X_i),\ldots,\sigma_L^2(X_i)\big)
\Big).
\end{split}
\label{eq:nll}
\end{equation}
which can be optimized end-to-end via backpropagation.
Regularization terms such as bounded-variance constraints, smoothness penalties on $a_j(X)$ and $c_j(X)$, or spatial coherence priors can be incorporated to improve numerical stability and interpretability in large-scale settings.

\paragraph{Interpretation.}
Maximizing the marginal likelihood encourages agreement between sensors after accounting for learned bias and reliability.
Sensors with larger learned variance receive smaller precision weights in the posterior aggregation and therefore exert less influence on the fused latent estimate, whereas sensors with higher precision, quantified by $a_j^2(X_i)/\sigma_j^2(X_i)$, contribute more strongly to the posterior distribution. This enables fully unsupervised learning of fusion parameters without access to ground-truth labels.

\paragraph{Variance Regularization.}

While maximizing the marginal likelihood in Eq.~\eqref{eq:nll}
provides a principled unsupervised objective,
it does not explicitly penalize excessively large learned variances.
In practice, we observed that the sensor noise variances
$\sigma_j^2(X_i)$ and prior variance $\sigma_0^2(X_i)$
may grow during training without substantially affecting the marginal likelihood, leading to unstable posterior estimates
and degraded predictive RMSE.

To mitigate this effect, we introduce an explicit variance
regularization term:
\begin{equation}
\mathcal{R}_{\mathrm{var}}
=
\lambda_0 \sum_{i=1}^N \left(\log \sigma_0^2(X_i)\right)^2
+
\lambda_s \sum_{i=1}^N \sum_{j=1}^L
\left(\log \sigma_j^2(X_i)\right)^2.
\label{eq:var_reg}
\end{equation}
where $\lambda_0$ and $\lambda_s$ are non-negative
regularization hyperparameters controlling the strength of
prior and sensor variance penalties. This regularization mitigates variance inflation degeneracies and encourages improvements in marginal likelihood to align with improvements in predictive accuracy, corresponding to penalized marginal likelihood estimation.

The total training objective becomes:
\begin{equation}
\mathcal{L}_{\mathrm{total}}
=
\mathcal{L}_{\mathrm{NCAM}}
+
\mathcal{R}_{\mathrm{var}}.
\label{eq:total_objective}
\end{equation}

\subsection{Fused Estimate and Uncertainty Quantification}

After training, the model provides for each entity $i$ a posterior distribution
over the latent variable $\Theta_i$ as well as a posterior predictive distribution
for the true (unobserved) target variable $Y_i$.

\paragraph{Latent Aggregation Output.}
The aggregation posterior yields
\begin{align}
\mathbb{E}[\Theta_i \mid B_i, X_i]
&= \mu_{\mathrm{post}}(B_i,X_i), \nonumber\\
\mathrm{Var}[\Theta_i \mid B_i, X_i]
&= \sigma_{\mathrm{post}}^2(B_i,X_i),
\label{eq:latent_fused}
\end{align}
which quantify the fused latent estimate and its associated epistemic uncertainty propagated through the multi-sensor aggregation and prior model.

\paragraph{Posterior Predictive Output.}
For downstream prediction, the relevant quantity is the posterior predictive
distribution
\[
p(Y_i \mid B_i, X_i)
=
\mathcal{N}\!\left(
Y_i;\,
\mu_{\mathrm{post}}(B_i,X_i),\,
\sigma_{\mathrm{post}}^2(B_i,X_i) + \sigma_y^2
\right).
\]

We define the fused predictive mean as
\begin{equation}
\hat{Y}_i^{\mathrm{fused}}
=
\mathbb{E}[Y_i \mid B_i, X_i]
=
\mu_{\mathrm{post}}(B_i,X_i),
\label{eq:fused_mean_out}
\end{equation}
and the total predictive uncertainty as
\begin{equation}
\hat{\sigma}_{i,\mathrm{pred}}^2
=
\mathrm{Var}(Y_i \mid B_i, X_i)
=
\sigma_{\mathrm{post}}^2(B_i,X_i) + \sigma_y^2.
\label{eq:fused_var_out}
\end{equation}

\paragraph{Credible Intervals.}
Model-based $(1-\alpha)$ credible intervals are obtained as
\begin{equation}
\hat{Y}_i^{\mathrm{fused}}
\;\pm\;
z_{1-\alpha/2}
\sqrt{\hat{\sigma}_{i,\mathrm{pred}}^2},
\label{eq:credible}
\end{equation}
where $z_{1-\alpha/2}$ denotes the standard normal quantile.

These intervals reflect epistemic uncertainty propagated through the multi-sensor aggregation and aleatoric uncertainty intrinsic to the underlying process. They serve as uncalibrated model-based uncertainty estimates that are subsequently refined via Monte Carlo Conformal Prediction.


\subsection{Monte Carlo Conformal Calibration}

Our Monte Carlo conformal calibration procedure is inspired by the Monte Carlo conformal framework of \citet{stutz2023ambiguous}, which introduces sampling-based conformal prediction under ambiguous ground truth. We adapt this algorithm to continuous regression with latent targets and integrate it with conjugate Bayesian aggregation, yielding a ground-truth–free uncertainty calibration mechanism for unsupervised multi-sensor fusion. For completeness, we provide the regression-adapted MC-CP procedure in Appendix~\ref{app:mc_cp_algorithm}.

While the Bayesian model provides principled predictive uncertainties,
these uncertainties may be miscalibrated relative to finite-sample observations.
We therefore employ a \emph{locally adaptive Monte Carlo Conformal Prediction (MC-CP)}
procedure that explicitly accounts for heteroscedastic uncertainty.

\paragraph{Monte Carlo Sampling.}
For each calibration entity $i$, synthetic realizations are generated from
the posterior predictive model using a two-stage sampling procedure:
\begin{align}
\Theta_i^{(r)} &\sim p(\Theta_i \mid B_i, X_i), \nonumber\\
Y_i^{(r)} &\sim p(Y_i \mid \Theta_i^{(r)}).
\label{eq:mcsampling}
\end{align}
This hierarchical sampling scheme is mathematically equivalent to drawing samples directly from the posterior predictive distribution
$p(Y_i \mid B_i, X_i)$ while making explicit the decomposition between epistemic and aleatoric uncertainty.

Unlike classical split conformal prediction, which relies on observed calibration targets, the proposed MC-CP procedure calibrates the posterior predictive distribution via Monte Carlo samples, thereby adjusting model-based uncertainty while preserving finite-sample coverage guarantees with respect to the calibrated predictive distribution under exchangeability.

\paragraph{Locally Adaptive Conformity Scores.}
To account for heteroscedastic predictive uncertainty,
we define \emph{variance-normalized conformity scores} as
\begin{equation}
s_i^{(r)}
=
\frac{
\big| Y_i^{(r)} - \hat{Y}_i^{\mathrm{fused}} \big|
}{
\hat{\sigma}_{i,\mathrm{pred}}
},
\label{eq:adaptive_scores}
\end{equation}
where
\[
\hat{Y}_i^{\mathrm{fused}} = \mathbb{E}[Y_i \mid B_i, X_i],
\qquad
\hat{\sigma}_{i,\mathrm{pred}}^2
=
\mathrm{Var}(Y_i \mid B_i, X_i).
\]

Let $q_{1-\alpha}$ denote the empirical $(1-\alpha)$ quantile
of the pooled normalized conformity scores $\{s_i^{(r)}\}$.

\paragraph{Calibrated Prediction Intervals.}
The locally adaptive conformal prediction interval for entity $i$ is given by
\begin{equation}
[L_i, U_i]
=
\left[
\hat{Y}_i^{\mathrm{fused}} - q_{1-\alpha}\,\hat{\sigma}_{i,\mathrm{pred}},
\;
\hat{Y}_i^{\mathrm{fused}} + q_{1-\alpha}\,\hat{\sigma}_{i,\mathrm{pred}}
\right],
\label{eq:adaptive_intervals}
\end{equation}
which achieves finite-sample coverage guarantees under exchangeability with respect to the conformalized posterior predictive distribution, and approximate coverage under weak temporal dependence induced by block splitting.

\paragraph{Outputs.}
Locally adaptive MC-CP produces:
(i) Bayesian fused point predictions $\hat{Y}_i^{\mathrm{fused}}$,
(ii) heteroscedastic, conformalized prediction intervals with guaranteed finite-sample coverage,
and
(iii) uncertainty-aware intervals whose width adapts to local predictive variance.

\subsection{Sensor-Anchored Conformal Calibration}

Monte Carlo Conformal Prediction (MC-CP) provides coverage guarantees with respect
to the conformalized posterior predictive distribution under exchangeability.
However, when the generative assumptions are misspecified,
MC-CP alone cannot detect inconsistency between the fused prediction and
the observed sensor measurements.
In this subsection, we introduce an additional
\emph{sensor-anchored conformal calibration} mechanism that directly grounds
the fused predictions in the observable data.

\paragraph{Motivation.}
While MC-CP calibrates uncertainty at the latent target level,
it relies on samples generated from the learned model.
As a result, a misspecified but internally consistent model
may still exhibit nominal coverage.
To address this limitation, we propose a complementary calibration procedure
that evaluates the consistency of the fused prediction with respect to the
observed sensor measurements themselves.

\paragraph{Bias-Corrected Sensor Residuals.}
Recall the sensor measurement model:
\begin{equation}
b_{ij} = a_j(X_i)\Theta_i + c_j(X_i) + \epsilon_{ij}.
\end{equation}
Using the learned sensor calibration parameters,
we define a bias-corrected sensor-implied latent estimate as
\begin{equation}
\tilde{\Theta}_{ij}
=
\frac{b_{ij} - c_j(X_i)}{a_j(X_i)}.
\end{equation}

The corresponding sensor-anchored residual is then given by
\begin{equation}
r_{ij}
=
\left|
\tilde{\Theta}_{ij}
-
\hat{Y}_i^{\mathrm{fused}}
\right|
=
\left|
\frac{b_{ij} - c_j(X_i)}{a_j(X_i)}
-
\hat{Y}_i^{\mathrm{fused}}
\right|.
\label{eq:sensor_residual}
\end{equation}

\paragraph{Sensor-Anchored Conformity Scores.}
To account for heteroscedastic uncertainty, we define
\emph{locally adaptive sensor-anchored conformity scores} as
\begin{equation}
s_{ij}^{\mathrm{sens}}
=
\frac{
\left|
\frac{b_{ij} - c_j(X_i)}{a_j(X_i)}
-
\hat{Y}_i^{\mathrm{fused}}
\right|
}{
\hat{\sigma}_{i,\mathrm{pred}}
},
\label{eq:sensor_adaptive_score}
\end{equation}
where $\hat{\sigma}_{i,\mathrm{pred}}^2 = \mathrm{Var}(Y_i \mid B_i, X_i)$
denotes the posterior predictive variance defined in Eq.~\eqref{eq:fused_var_out}.

Let $q^{\mathrm{sens}}_{1-\alpha}$ denote the empirical $(1-\alpha)$ quantile
of the pooled sensor-anchored conformity scores
$\{s_{ij}^{\mathrm{sens}}\}_{i,j}$.

\paragraph{Sensor-Anchored Prediction Interval.}
A locally adaptive sensor-anchored conformal interval is then given by
\begin{equation}
I_i^{\mathrm{sens}}
=
\left[
\hat{Y}_i^{\mathrm{fused}}
-
q^{\mathrm{sens}}_{1-\alpha}\,\hat{\sigma}_{i,\mathrm{pred}},
\;
\hat{Y}_i^{\mathrm{fused}}
+
q^{\mathrm{sens}}_{1-\alpha}\,\hat{\sigma}_{i,\mathrm{pred}}
\right].
\label{eq:sensor_interval}
\end{equation}
\paragraph{Interpretation.}
The interval in Eq.~\eqref{eq:sensor_interval} quantifies the degree to which
the fused prediction is consistent with the bias-corrected sensor observations.
Unlike posterior-sampling-based MC-CP, this calibration is grounded directly in observed data and can
expose systematic model misspecification that remains undetected by
model-based calibration alone.

\paragraph{Complementarity with MC-CP.}
The proposed sensor-anchored calibration does not replace MC-CP,
but rather complements it.
MC-CP provides calibrated uncertainty with respect to the model-implied predictive distribution,
while sensor-anchored calibration provides an external consistency check
against observable measurements.
Differences between the two intervals may indicate model misspecification
or limitations in the learned uncertainty representation.

\subsection{Temporal Block Splitting and Exchangeability}
\label{sec:temporal_split}

Environmental sensor data exhibit strong temporal dependence and seasonal structure, so naive random splits violate exchangeability and induce leakage across training, calibration, and test sets. We therefore adopt a block-based temporal splitting strategy: the timeline is partitioned into contiguous blocks of length $B$ (7 days), separated by a temporal gap $G$ (36 hours), and entire blocks are assigned to train/validation/calibration/test splits. This reduces short-range leakage while allowing the model to observe patterns from across the year.

The 36-hour gap is selected via residual autocorrelation analysis, choosing the smallest lag at which the ACF of model residuals falls within the 95\% confidence band (Appendix~\ref{app:acf_gap_selection}).

Classical conformal guarantees require exchangeability, which is violated in time-series data. By separating blocks with a gap, we reduce cross-split dependence and interpret conformal coverage as approximate under weak temporal dependence. We report empirical coverage on held-out test data to quantify any residual miscalibration.

\begin{table*}[t]
  \centering
  \caption{Baseline Metrics Across Datasets (Best Single-Sensor per Dataset). Deltas are \% worse than Anchored NCAM (lower is better). Best is bold; second-best is underlined.}\label{tab:baseline-metrics-merged}
  \setlength{\tabcolsep}{5pt}
  \renewcommand{\arraystretch}{1.05}
  \begin{tabular}{lrrrrrr}
    \toprule
    \bfseries Method
    & \multicolumn{2}{c}{\bfseries Toy}
    & \multicolumn{2}{c}{\bfseries SensEURCity}
    & \multicolumn{2}{c}{\bfseries CAIRSENSE} \\
    \cmidrule(lr){2-3}\cmidrule(lr){4-5}\cmidrule(lr){6-7}
    & \bfseries RMSE ($\Delta$\%) & \bfseries MAE ($\Delta$\%)
    & \bfseries RMSE ($\Delta$\%) & \bfseries MAE ($\Delta$\%)
    & \bfseries RMSE ($\Delta$\%) & \bfseries MAE ($\Delta$\%) \\
    \midrule
    Anchored NCAM
      & \textbf{1.754} {\scriptsize (+0\%)} & \textbf{1.398} {\scriptsize (+0\%)} & \textbf{5.454} {\scriptsize (+0\%)} & \textbf{3.490} {\scriptsize (+0\%)} & \textbf{2.566} {\scriptsize (+0\%)} & \underline{1.988} {\scriptsize (+0\%)} \\
    Kalman Filter
      & 3.574 {\scriptsize (+104\%)} & 3.104 {\scriptsize (+122\%)} & 7.727 {\scriptsize (+42\%)} & 4.790 {\scriptsize (+37\%)} & 2.661 {\scriptsize (+4\%)} & 2.036 {\scriptsize (+2\%)} \\
    Probabilistic PCA
      & 4.629 {\scriptsize (+164\%)} & 4.212 {\scriptsize (+201\%)} & 9.146 {\scriptsize (+68\%)} & 5.362 {\scriptsize (+54\%)} & 2.755 {\scriptsize (+7\%)} & 2.098 {\scriptsize (+6\%)} \\
    Mean Aggregation
      & 5.023 {\scriptsize (+186\%)} & 4.337 {\scriptsize (+210\%)} & 7.206 {\scriptsize (+32\%)} & 4.599 {\scriptsize (+32\%)} & \underline{2.595} {\scriptsize (+1\%)} & \textbf{1.974} {\scriptsize (-1\%)} \\
    Inverse-Variance Weighting
      & 3.348 {\scriptsize (+91\%)} & 2.733 {\scriptsize (+96\%)} & 8.380 {\scriptsize (+54\%)} & 5.461 {\scriptsize (+56\%)} & 2.761 {\scriptsize (+8\%)} & 2.121 {\scriptsize (+7\%)} \\
    VELI~\citep{dalbah2025veli}
      & \underline{2.644} {\scriptsize (+51\%)} & \underline{2.117} {\scriptsize (+51\%)} & \underline{6.176} {\scriptsize (+13\%)} & \underline{4.228} {\scriptsize (+21\%)} & 3.277 {\scriptsize (+28\%)} & 2.469 {\scriptsize (+24\%)} \\
    \bottomrule
  \end{tabular}
\end{table*}

\section{Experiments}\label{sec:experiments}

We evaluate NCAM on one synthetic and two real-world datasets spanning spatial, temporal, and spatio-temporal settings. All models are trained without ground-truth supervision; reference labels are used only for evaluation.

\subsection{Experimental Setup}\label{subsec:setup}

\paragraph{Datasets}
We cover three types of datasets: spatial, temporal, and spatio-temporal. Because the sensor structures differ across datasets, we use dataset-specific preprocessing and splits. Due to heavy seasonality in PM$_{2.5}$ and the short span of some sensor records, we also use different split strategies across datasets.
\begin{itemize}
    \item \textbf{Toy.} A synthetic 2D grid with known ground truth. We generate three sensors with distinct multiplicative and additive biases and Gaussian noise, and use a random split. The covariates are four periodic spatial features: sinusoidal components of latitude and longitude (sine/cosine terms). Exact data creation steps are given in Appendix~\ref{app:toy-data-generation}.
    
    \item \textbf{SensEURCity.} \citet{senseurcity} A spatio-temporal PM$_{2.5}$ dataset with three low-cost sensor readings from Antwerp, Oslo, and Zagreb. We use a chronological split (60\% train, 10\% validation, 15\% calibration, 15\% test) with 36-hour gaps between splits to mitigate temporal leakage. Data are mean-aggregated hourly. We only use Antwerp because of the heavy distributional shift across cities. Highly correlated duplicate sensors were removed: we retained 5325CAT and OPCN3PM25 and dropped 5325CST due to its high correlation with 5325CAT. The covariates include location (latitude and longitude), temperature, humidity-related variables, dew-point deficit, and air pressure.
    
    \item \textbf{CAIRSENSE.} \citet{cairsense} A co-located PM$_{2.5}$ sensor network using AirAssure sensors. Because the sensors are co-located and exhibit strong short-range temporal dependence, we use block-based temporal splitting with 7-day blocks and a 36-hour gap between blocks; blocks are assigned to train/val/cal/test in the same 60/10/15/15 proportions. We select the gap length by inspecting residual autocorrelation at short lags and choosing the smallest gap for which autocorrelation falls within the 95\% noise band (see Appendix~\ref{app:acf_gap_selection}). The covariates include temperature, relative humidity, wind speed, and wind direction, plus sinusoidal time features for hour of day, day of year, and day of week.
\end{itemize}

\paragraph{Baselines and Metrics.}
Since NCAM trains without ground truth, we restrict comparisons to unsupervised baselines: mean aggregation, inverse-variance weighting, probabilistic PCA, Kalman filtering, and VELI~\citep{dalbah2025veli}, a recent variational latent-variable model for unsupervised low-cost-sensor calibration (we report its stronger anchored-readout variant).
Point prediction quality is measured by RMSE and MAE against the held-out reference sensor. For uncertainty quantification, we report empirical coverage and average interval width of conformal prediction intervals at target level $1-\alpha=0.9$.

\paragraph{Implementation Details.}
All probabilistic components are parameterized by small MLP networks with three hidden layers, GELU activations, and a linear output layer. The model uses three heads: a prior head outputting $(\mu_0(X_i), \log \sigma_0^2(X_i))$, a reliability head outputting per-sensor log-variances $\{\log \sigma_j^2(X_i)\}_{j=1}^L$, and a bias head outputting multiplicative and additive calibration terms. In the anchored variant, the anchor sensor is fixed to scale $1$ and offset $0$. Log-variance values are clamped to $[-5,4]$ to prevent collapse or explosion, and regularization terms penalize extreme prior-variance and sensor-variance predictions. We use the same regularization coefficient for both prior and sensor variance penalties. This coefficient is set to $0.0$ for the Toy dataset and $1.0$ for SensEURCity and CAIRSENSE. We evaluated multiple aleatoric noise levels and observed only marginal effects on performance; accordingly, we retained $10^{-3}$ as the default across experiments.


\subsection{Predictive Performance}\label{subsec:predictive}

Table~\ref{tab:baseline-metrics-merged} summarizes point prediction results across all three datasets. Anchored NCAM achieves the best or near-best performance on every dataset-metric combination.

On the \textbf{Toy dataset}, where sensors have known heterogeneous multiplicative and additive biases, NCAM achieves RMSE~1.754 and MAE~1.398. The strongest baseline is VELI, which is 51\% worse on both RMSE and MAE; the best classical baseline, inverse-variance weighting, is 91\% worse on RMSE and 96\% worse on MAE. Mean aggregation degrades by 186\%/210\%, and Kalman filtering by 104\%/122\%. The large margins confirm that explicit bias correction is essential when sensors are heterogeneously corrupted; methods that assume unbiased or equally reliable sources cannot compensate.

On \textbf{SensEURCity}, NCAM achieves RMSE~5.454 and MAE~3.490. The second-best method is VELI, 13\% worse on RMSE and 21\% worse on MAE; the best classical baseline, mean aggregation, is 32\% worse on both metrics, and the Kalman filter is 42\%/37\% worse. The gains are smaller than on the Toy data but still substantial, reflecting moderate sensor bias. Even without explicit sensor identity information, the model learns spatially varying reliability through covariates.

On \textbf{CAIRSENSE}, the classical baselines cluster within an 8\% range. NCAM leads on RMSE (2.566, best by 1\%) but mean aggregation slightly outperforms on MAE (1.974 vs.\ 1.988, a~1\% margin); VELI is less competitive in this regime ($+28\%$/$+24\%$). This is consistent with the co-located setup: sensors measuring the same air mass exhibit small biases, and simple averaging is competitive. Importantly, NCAM does not overfit in this low-bias regime despite its additional model capacity.

Overall, NCAM's advantage scales with the severity of inter-sensor bias, delivering the largest gains where multiplicative and additive corruption is significant while remaining competitive in the nearly homogeneous CAIRSENSE setting.

\begin{table*}[thb]
  \centering
  \caption{Conformal Metrics Across Datasets.}\label{tab:conformal-metrics-merged}
  \small
  \setlength{\tabcolsep}{4pt}
  \renewcommand{\arraystretch}{1.0}
  \resizebox{\textwidth}{!}{
  \begin{tabular}{lrrrrrr}
    \toprule
    \bfseries Method
    & \multicolumn{2}{c}{\bfseries Toy}
    & \multicolumn{2}{c}{\bfseries SensEURCity}
    & \multicolumn{2}{c}{\bfseries CAIRSENSE} \\
    \cmidrule(lr){2-3}\cmidrule(lr){4-5}\cmidrule(lr){6-7}
    & \bfseries Coverage & \bfseries Width
    & \bfseries Coverage & \bfseries Width
    & \bfseries Coverage & \bfseries Width \\
    \midrule
    Split Conformal
      & $0.901 \pm 0.003$ & $5.798 \pm 0.026$ & $0.900 \pm 0.002$ & $20.808 \pm 0.135$ & $0.903 \pm 0.061$ & $40.725 \pm 36.691$ \\
    Fused Posterior Predictive
      & $0.889 \pm 0.002$ & $5.606 \pm 0.000577$ & $0.973 \pm 0.000659$ & $34.839 \pm 0.058$ & $0.808 \pm 0.035$ & $10.445 \pm 0.798$ \\
    \midrule
    MC-CP Empirical
      & $1.000 \pm 0.000$ & $23.185 \pm 0.071$ & $0.977 \pm 0.000654$ & $37.119 \pm 0.091$ & $0.897 \pm 0.062$ & $30.996 \pm 17.520$ \\
    MC-CP (Model)
      & $0.888 \pm 0.002$ & $5.593 \pm 0.000576$ & $0.972 \pm 0.000658$ & $34.785 \pm 0.058$ & $0.808 \pm 0.035$ & $10.456 \pm 0.799$ \\
    MC-CP Sensor-Anchored
      & $0.994 \pm 0.000427$ & $9.824 \pm 0.035$ & $0.997 \pm 0.000267$ & $68.786 \pm 0.520$ & $0.808 \pm 0.098$ & $12.376 \pm 5.062$ \\
    \bottomrule
  \end{tabular}
  }
\end{table*}

\subsection{Non-Identifiability and Sensor Anchoring}\label{subsec:identifiability}

Without anchoring, structural non-identifiability allows the model to reduce NLL by absorbing signal into bias parameters, causing the posterior mean to drift. As shown in Fig.~\ref{fig:anchored-unanchored} (Appendix~\ref{app:identifiability}), the validation NLL decreases while MAE increases, indicating improved likelihood fit without corresponding gains in predictive accuracy.

Anchoring one sensor by fixing its multiplicative bias to 1 and additive bias to 0 breaks this symmetry and resolves the pathology. Under anchoring, NLL and MAE decrease together, reflecting stable optimization and meaningful predictive improvement. On the Toy dataset, the learned sensor noise variances under the anchored model follow the ground-truth reliability ordering ($\sigma_1^2 < \sigma_2^2 < \sigma_3^2$) consistently across 10 independent runs (Spearman $\rho=0.94$, bootstrap 95\% CI $[0.91, 0.97]$), confirming that the model captures true inter-sensor reliability structure.

\begin{table}[b]
    \centering
    \caption{Ablation of multiplicative and additive bias (delta w.r.t.\ Anch.)}
    \label{tab:multiplicative_additive_ablation}
    \small
    \setlength{\tabcolsep}{3.5pt}
    \renewcommand{\arraystretch}{1.0}
    \begin{tabular}{@{}lcccc@{}}
        \toprule
        \bfseries Dataset & \multicolumn{2}{c}{\bfseries RMSE} & \multicolumn{2}{c}{\bfseries MAE} \\
        \cmidrule(lr){2-3}\cmidrule(lr){4-5}
        & \bfseries Anch. & \bfseries No-Bias ($\Delta$\%)
        & \bfseries Anch. & \bfseries No-Bias ($\Delta$\%) \\
        \midrule
        Toy Dataset & \textbf{1.754} & 3.206 (+82.8\%) & \textbf{1.398} & 2.675 (+91.4\%) \\
        SenseurCity & \textbf{5.454} & 9.726 (+78.3\%) & \textbf{3.490} & 6.218 (+78.1\%) \\
        CAIRSENSE & 2.566 & \textbf{2.446} (-4.7\%) & 1.988 & \textbf{1.882} (-5.3\%) \\
        \bottomrule
    \end{tabular}
\end{table}

\subsection{Bias Modeling Ablation}\label{subsec:ablation}

To isolate the contribution of explicit bias modeling, we compare the full anchored NCAM against a variant with both multiplicative and additive bias heads removed (Table~\ref{tab:multiplicative_additive_ablation}). Removing bias modeling degrades performance dramatically on Toy (+82.8\% RMSE, +91.4\% MAE) and SensEURCity (+78.3\% RMSE, +78.1\% MAE), confirming that bias correction is essential when sensor relationships are weaker and sensors exhibit heterogeneous error characteristics. This is particularly expected on SensEURCity, where sensors are of different types and show lower inter-sensor correlation than CAIRSENSE. In contrast, on CAIRSENSE, the no-bias variant is slightly better ($-4.7$\% RMSE, $-5.3$\% MAE), consistent with its co-located, same-type, highly correlated sensor setup: shared signal structure already aligns sensors well, leaving limited systematic bias to correct. This ablation validates the architectural choice of separate bias heads and explains the dataset dependent performance profile observed in Table~\ref{tab:baseline-metrics-merged}.

\subsection{Conformal Prediction Intervals}\label{subsec:conformal}

We apply several conformal prediction methods that differ only in the choice of calibration target $y_i$. \textbf{Split conformal:} uses the ground-truth label from the calibration split. \textbf{MC-CP (Model):} uses a pseudo-label sampled from the posterior predictive distribution. \textbf{MC-CP Empirical:} uses a pseudo-label sampled from the empirical sensor distribution. \textbf{MC-CP Sensor-Anchored:} uses a bias-corrected sensor reading as a proxy target.

We evaluate empirical coverage and average interval width at target level $1-\alpha=0.9$ (Table~\ref{tab:conformal-metrics-merged}). Split conformal is included as an idealized reference requiring ground-truth labels for calibration, which are typically unavailable in our deployment setting. In contrast, MC-CP~Empirical, MC-CP~(Model), and MC-CP~Sensor-Anchored operate without target-domain labels. 

Notably, the Fused Posterior Predictive row as calculated using Eq.~\ref{eq:credible} and MC-CP (Model) rows yield nearly identical coverage and interval width. This is expected: because MC-CP (Model) samples from and calibrates using the same Gaussian posterior predictive distribution, it effectively recovers the model-implied quantile. The empirical agreement therefore verifies that the Monte Carlo calibration behaves as expected relative to the underlying Bayesian predictive distribution.

Split conformal achieves near-nominal coverage across datasets with moderate widths. Among ground-truth-free methods, MC-CP~Empirical consistently over-covers with substantially wider intervals, indicating conservative uncertainty estimates driven by empirical sensor variability. MC-CP~(Model) yields the tightest intervals by leveraging the model's locally adaptive predictive variance, but exhibits under-coverage in regimes with heavier-tailed residuals or increased dispersion. MC-CP~Sensor-Anchored behaves similarly to the empirical variant on some datasets while under-covering in others, reflecting sensitivity to sensor-level dispersion.

Overall, the results reveal a clear coverage--width tradeoff. MC-CP~(Model) provides the most interval-efficient solution when coverage remains acceptable, whereas MC-CP~Empirical and MC-CP~Sensor-Anchored offer more conservative uncertainty at the expense of interval width. The large variability in interval width observed on CAIRSENSE suggests heavy-tailed residual structure, which challenges all conformal variants and highlights the importance of adaptive uncertainty modeling.

\section{Additional Experiments}\label{app:additional_experiments}

This section reports four supporting studies probing the robustness and cost of the method: sensitivity to hyperparameter and architecture choices, data efficiency under reduced training fractions, sensitivity to the choice of anchor sensor, and the effect of the Monte Carlo sample count on MC-CP calibration. Results are reported as means over the indicated number of seeds; CAIRSENSE follows the same last-fold temporal protocol as the main experiments (Section~\ref{sec:experiments}).

\subsection{Hyperparameter and Architecture Sensitivity}\label{app:hp_sensitivity}

We vary the network width (hidden dimension) and the learning rate independently around each dataset's default configuration, holding all other settings fixed, and repeat every configuration over $10$ seeds. Table~\ref{tab:hp_sensitivity} reports the resulting span of test RMSE across each grid. On the Toy data the span is below $1\%$ for both axes, comparable to seed-to-seed noise. On the real datasets it stays bounded at most $11.9\%$ across the five hidden-dimension settings and $14.8\%$ across the five learning rates indicating that NCAM does not require finely tuned capacity or step size to reach competitive accuracy.

\begin{table*}[t]
  \centering
  \caption{Hyperparameter and architecture sensitivity. Test RMSE range (minimum--maximum) and relative span across a five-point hidden-dimension grid and a five-point learning-rate grid, each averaged over $10$ seeds. Hidden-dimension grid: $\{32,64,128,256,512\}$ (Toy), $\{2,4,8,16,32\}$ (SensEURCity, CAIRSENSE); learning-rate grid: $\{1,2,4,10,20\}\times10^{-5}$ (Toy), $\{5,10,20,50,100\}\times10^{-5}$ (SensEURCity, CAIRSENSE).}
  \label{tab:hp_sensitivity}
  \setlength{\tabcolsep}{6pt}
  \renewcommand{\arraystretch}{1.1}
  \begin{tabular}{lcc}
    \toprule
    \bfseries Dataset & \bfseries Hidden dim (RMSE span) & \bfseries Learning rate (RMSE span) \\
    \midrule
    Toy          & 1.772--1.787 \,(0.9\%)  & 1.772--1.779 \,(0.4\%) \\
    SensEURCity  & 5.509--6.166 \,(11.9\%) & 5.452--5.660 \,(3.8\%) \\
    CAIRSENSE    & 2.531--2.703 \,(6.8\%)  & 2.532--2.906 \,(14.8\%) \\
    \bottomrule
  \end{tabular}
\end{table*}

\subsection{Data Efficiency}\label{app:data_efficiency}

We retrain NCAM on random subsamples of the training split at $\{100, 50, 25, 12.5\}\%$ of its size, keeping the validation, calibration, and test splits fixed, and average over $6$ seeds (Table~\ref{tab:data_efficiency}). On Toy and SensEURCity, accuracy is essentially flat down to $25\%$ of the data and degrades only mildly at $12.5\%$, and NCAM stays well ahead of the best classical baseline in Table~\ref{tab:baseline-metrics-merged} at every fraction. On CAIRSENSE the error rises gently and monotonically as data is removed (a $14\%$ RMSE increase from full to one-eighth); in this near-homogeneous, low-bias regime NCAM tracks the strongest classical baseline (mean aggregation) rather than overtaking it, consistent with the main-table result. The $100\%$ row is a $6$-seed mean and therefore differs slightly from the single-run values in Table~\ref{tab:baseline-metrics-merged}.

\begin{table*}[!t]
  \centering
  \caption{Data efficiency: NCAM test RMSE / MAE at reduced training fractions, averaged over $6$ seeds.}
  \label{tab:data_efficiency}
  \setlength{\tabcolsep}{5pt}
  \renewcommand{\arraystretch}{1.1}
  \begin{tabular}{lcccccc}
    \toprule
    & \multicolumn{2}{c}{\bfseries Toy} & \multicolumn{2}{c}{\bfseries SensEURCity} & \multicolumn{2}{c}{\bfseries CAIRSENSE} \\
    \cmidrule(lr){2-3}\cmidrule(lr){4-5}\cmidrule(lr){6-7}
    \bfseries Train frac. & RMSE & MAE & RMSE & MAE & RMSE & MAE \\
    \midrule
    $100\%$   & 1.774 & 1.412 & 5.650 & 3.605 & 2.636 & 2.020 \\
    $50\%$    & 1.772 & 1.411 & 5.677 & 3.626 & 2.804 & 2.112 \\
    $25\%$    & 1.782 & 1.418 & 5.719 & 3.660 & 2.912 & 2.181 \\
    $12.5\%$  & 1.816 & 1.442 & 6.491 & 4.265 & 3.003 & 2.238 \\
    \bottomrule
  \end{tabular}
\end{table*}

\subsection{Sensitivity to Anchor Selection}\label{app:anchor_sensitivity}

\begin{table}[!t]
  \centering
  \caption{Sensitivity to anchor selection. NCAM test RMSE / MAE when anchored on each sensor ($6$ seeds), and the corresponding raw single-sensor readings against the reference. $\dagger$ marks the anchor used in the main experiments (Table~\ref{tab:baseline-metrics-merged}).}
  \label{tab:anchor_sensitivity}
  \scriptsize
  \setlength{\tabcolsep}{2.2pt}
  \renewcommand{\arraystretch}{1.0}
  \resizebox{\columnwidth}{!}{%
  \begin{tabular}{llcccc}
    \toprule
    & & \multicolumn{2}{c}{\bfseries NCAM (anchored)} & \multicolumn{2}{c}{\bfseries Raw sensor} \\
    \cmidrule(lr){3-4}\cmidrule(lr){5-6}
    \bfseries Dataset & \bfseries Anchor & RMSE & MAE & RMSE & MAE \\
    \midrule
    Toy & sensor\_0$^\dagger$ & 1.774  & 1.412  & 2.434  & 1.939 \\
        & sensor\_1           & 3.033  & 2.558  & 4.691  & 3.757 \\
        & sensor\_2           & 10.297 & 9.851  & 12.032 & 10.254 \\
    \cmidrule(lr){1-6}
    SensEURCity & OPCN3PM25         & 10.212 & 6.658  & 11.740 & 8.268 \\
                & 5325CAT$^\dagger$ & 5.650  & 3.605  & 7.442  & 5.074 \\
    \cmidrule(lr){1-6}
    CAIRSENSE & AirAssure1            & 2.704 & 2.061 & 4.516 & 3.888 \\
              & AirAssure2            & 2.546 & 1.949 & 2.871 & 2.210 \\
              & AirAssure3$^\dagger$  & 2.636 & 2.020 & 2.470 & 1.988 \\
    \bottomrule
  \end{tabular}%
  }
\end{table}

The anchored parameterization (Appendix~\ref{app:identifiability}) fixes one sensor's affine calibration to resolve the latent scale/offset non-identifiability. Table~\ref{tab:anchor_sensitivity} reports NCAM accuracy when each sensor in turn is used as the anchor ($6$ seeds per anchor), alongside that sensor's raw readings against the reference. When sensors are heterogeneous, the anchor matters: on Toy, anchoring on the least reliable sensor inflates RMSE by roughly $5\times$ relative to the best anchor, and on SensEURCity the better anchor more than halves the error. When sensors are co-located and near-homogeneous (CAIRSENSE), accuracy is largely insensitive to the anchor (within $\sim$6\%). In every case except CAIRSENSE's single strongest sensor (AirAssure3, the most accurate raw sensor in the study), anchoring on a sensor yields lower error than that sensor's raw readings, confirming that NCAM corrects rather than merely echoes the anchor.

\subsection{MC-CP Sample-Count Sensitivity}\label{app:mc_sample_sensitivity}

\begin{table}[!t]
  \centering
  \caption{MC-CP sample-count sensitivity on Toy: empirical coverage (target $0.90$), mean interval width, and calibration wall-clock time as a function of the pseudo-label sample count $m$.}
  \label{tab:mc_sample_sensitivity}
  \footnotesize
  \setlength{\tabcolsep}{4.5pt}
  \renewcommand{\arraystretch}{1.0}
  \begin{tabular}{rccc}
    \toprule
    \bfseries $m$ & \bfseries Coverage & \bfseries Width & \bfseries Cal.\ time (s) \\
    \midrule
    1   & 0.8915 & 5.613 & 0.026 \\
    5   & 0.8931 & 5.646 & 0.023 \\
    10  & 0.8921 & 5.627 & 0.028 \\
    50  & 0.8910 & 5.603 & 0.049 \\
    100 & 0.8909 & 5.600 & 0.073 \\
    500 & 0.8909 & 5.601 & 0.343 \\
    \bottomrule
  \end{tabular}
\end{table}

We vary the number of Monte Carlo pseudo-label samples $m$ used by MC-CP (Appendix~\ref{app:mc_cp_algorithm}) on the Toy dataset, holding the trained model and the calibration split fixed. Table~\ref{tab:mc_sample_sensitivity} shows that empirical coverage and interval width are effectively constant across $m$ from $1$ to $500$: coverage stays within $0.891 \pm 0.001$ of the $0.90$ target and width within $\pm 0.05$. Calibration time grows roughly linearly with $m$ but remains a small fraction of a second even at $m=500$. A modest $m$ (e.g.\ $m \le 50$) therefore suffices, and the $m=50$ default used elsewhere incurs negligible overhead.

\FloatBarrier

\section{Conclusion}
We demonstrate that reliable multi-sensor regression is achievable without ground-truth supervision by combining principled probabilistic aggregation with uncertainty calibration. Across synthetic and real-world air-quality datasets, the method improves predictive accuracy in biased sensor regimes while remaining competitive when sensors are nearly homogeneous. Locally adaptive Monte Carlo conformal calibration provides practical heteroscedastic uncertainty intervals without target-domain labels, while sensor-anchored calibration offers a complementary measurement-grounded perspective. These results indicate that accurate and uncertainty-aware sensor fusion is feasible in fully unsupervised settings. Future work includes exploring dependence-aware calibration strategies and extending the framework to multivariate targets.

\newpage
\bibliography{uai2026-template}

\newpage

\onecolumn

\title{Supplementary Material}
\maketitle
\appendix

\section{Preliminary}\label{sec:preliminary}
\subsection{Conformal Prediction}
Conformal prediction is a framework for constructing prediction sets with
finite-sample, distribution-free coverage guarantees. In regression, the goal is
to produce an interval $\widehat{C}_\alpha(x)\subset\mathbb{R}$ such that for a
new test point $(X_{\text{new}},Y_{\text{new}})$,
\begin{equation}
\mathbb{P}\!\big(Y_{\text{new}}\in \widehat{C}_\alpha(X_{\text{new}})\big)\ge 1-\alpha,
\label{eq:cp-coverage}
\end{equation} 
under the assumption that the data points are exchangeable (e.g., i.i.d.).

\subsection{Locally Adaptive Conformal Prediction}
When predictive uncertainty is heteroscedastic, it is often beneficial to use a
\emph{normalized} nonconformity score. Suppose the model outputs a scale
estimate $\hat{\sigma}_{\mathrm{pred}}(x)>0$ (e.g., the predictive standard
deviation derived from a probabilistic model). Define the locally adaptive score
\begin{equation}
s_i = \frac{|y_i-\hat{\mu}(x_i)|}{\hat{\sigma}_{\mathrm{pred}}(x_i)},\qquad i\in\mathcal{I}_{\mathrm{cal}}.
\label{eq:local-adaptive-score}
\end{equation}
Let $q_{1-\alpha}$ be the empirical $(1-\alpha)$ quantile of these normalized
scores. The resulting locally adaptive conformal interval rescales by the test
point's predicted uncertainty:
\begin{equation}
\widehat{C}_\alpha(x) =
\big[\hat{\mu}(x)-q_{1-\alpha}\hat{\sigma}_{\mathrm{pred}}(x),\
\hat{\mu}(x)+q_{1-\alpha}\hat{\sigma}_{\mathrm{pred}}(x)\big].
\label{eq:local-adaptive-interval}
\end{equation}
This preserves the same marginal coverage guarantee as split conformal under exchangeability, while
allowing interval widths to vary across $x$.

\subsection{Adapted Monte Carlo Conformal Algorithm}\label{app:mc_cp_algorithm}
Monte Carlo conformal prediction (MC-CP) procedure introduced by \citet{stutz2023ambiguous} to construct prediction sets when calibration labels are ambiguous and represented by class plausibilities instead of a single hard label. The key idea is to sample multiple pseudo-labels from each plausibility vector and calibrate on the resulting expanded calibration set.

\paragraph{Setup.}
Let the calibration data be
\[
\{(X_i,\lambda_i)\}_{i=1}^n,
\]
where \(X_i\) is an input and
\[
\lambda_i = (\lambda_{i1},\ldots,\lambda_{iK})
\]
is a plausibility vector over \(K\) classes satisfying \(\lambda_{ik}\geq 0\) and \(\sum_{k=1}^K \lambda_{ik}=1\). Intuitively, \(\lambda_i\) encodes ambiguity in the ground-truth label for \(X_i\), for example after aggregating multiple annotator labels \citep{stutz2023ambiguous}.

Let \(E(x,k)\) denote a conformity score for class \(k\in\{1,\ldots,K\}\) at input \(x\). As discussed by \citet{stutz2023ambiguous}, a common choice is the predicted class probability, i.e., \(E(x,k)=\pi_k(x)\).

\paragraph{Monte Carlo Calibration Procedure.}
For each calibration example \(X_i\), MC-CP samples \(m\) pseudo-labels independently from the plausibility vector:
\begin{equation}
Y_i^1,\ldots,Y_i^m \sim \lambda_i.
\end{equation}
This produces an expanded calibration collection of size \(mn\),
\[
\{(X_i,Y_i^j)\}_{i=1,\ldots,n;\,j=1,\ldots,m},
\]
with conformity scores
\[
\{E(X_i,Y_i^j)\}_{i=1,\ldots,n;\,j=1,\ldots,m}.
\]

Let \(Q(\mathcal{S};q)\) denote the \(q\)-quantile of a finite set (or multiset) \(\mathcal{S}\). Algorithm~1 of \citet{stutz2023ambiguous} sets the calibration threshold to
\begin{equation}
\label{eq:mccp-threshold}
\tau
=
Q\!\left(
\left\{E(X_i,Y_i^j)\right\}_{i=1,\ldots,n;\,j=1,\ldots,m};
\frac{\lfloor \alpha m(n+1)\rfloor - m + 1}{mn}
\right),
\end{equation}
where \(\alpha\in(0,1)\) is the target miscoverage level.

The prediction set for a test input \(X\) is then
\begin{equation}
\label{eq:mccp-set}
C(X)=\{k\in\{1,\ldots,K\}: E(X,k)\geq \tau\}.
\end{equation}

\paragraph{Why MC-CP Differs From Majority-Vote Calibration.}
If calibration is performed using a single voted label (e.g., majority vote), label ambiguity is collapsed into a one-hot target. This can underestimate uncertainty when multiple classes are plausible. In contrast, MC-CP preserves ambiguity during calibration by sampling pseudo-labels from \(\lambda_i\), so calibration targets the aggregated label distribution rather than a hard vote \citep{stutz2023ambiguous}.

\paragraph{Coverage Interpretation.}
The relevant notion of coverage is coverage with respect to the aggregated label distribution:
\begin{equation}
\label{eq:agg-coverage}
\mathrm{P}_{\mathrm{agg}}(Y\in C(X))
=
\mathrm{E}_{X}\!\left[
\mathrm{E}_{Y\sim \mathrm{P}_{\mathrm{agg}}(\cdot\mid X)}
\bigl[\mathbf{1}\{Y\in C(X)\}\bigr]
\right].
\end{equation}
That is, coverage is averaged over both the input distribution and the label ambiguity encoded by the aggregated-label model \citep{stutz2023ambiguous}.

\paragraph{Guarantees and Practical Interpretation.}
\citet{stutz2023ambiguous} show that:
\begin{itemize}
    \item when \(m=1\), MC-CP reduces to standard split conformal prediction with labels sampled from the aggregated-label distribution and achieves the usual \(1-\alpha\) marginal coverage guarantee;
    \item when \(m\geq 2\), the repeated use of the same \(X_i\) with multiple sampled labels introduces dependence in the enlarged calibration set, so the standard exchangeability argument no longer applies directly; in this case, a conservative \(1-2\alpha\) guarantee is proved, while empirical coverage is often close to \(1-\alpha\).
\end{itemize}
In our NCAM-based regression framework, empirical coverage varies across datasets, reflecting model misspecification and temporal dependence; this behavior is consistent with the conservative $1-2\alpha$ guarantee for $m>1$ and the approximate exchangeability assumptions discussed above.

\section{NCAM Posterior and Predictive Derivations}
\subsection{Aggregation Posterior Derivation}
\label{sec:agg_posterior}

We derive the closed-form posterior distribution of the latent variable $\Theta_i$
given the observed multi-source measurements $B_i$ and covariates $X_i$.

Under the generative hierarchy
\begin{equation}
X_i \rightarrow \Theta_i \rightarrow (B_i, Y_i),
\end{equation}
the joint distribution factorizes as
\begin{equation}
\begin{split}
p(X_i, \Theta_i, B_i, Y_i)
&= p(X_i)\, p(\Theta_i \mid X_i) \\
&\quad \times p(B_i \mid \Theta_i, X_i)\, p(Y_i \mid \Theta_i).
\end{split}
\end{equation}

The aggregation task is to compute the latent posterior
\begin{equation}
p(\Theta_i \mid B_i, X_i)
=
\frac{p(B_i \mid \Theta_i, X_i)\, p(\Theta_i \mid X_i)}
     {p(B_i \mid X_i)}.
\label{eq:aggposterior}
\end{equation}

\paragraph{Likelihood Term.}
Assuming conditional independence of the measurements given $\Theta_i$,
the likelihood decomposes as
\begin{equation}
p(B_i \mid \Theta_i, X_i)
=
\prod_{j=1}^L
\mathcal{N}\!\left(
b_{ij};\, a_j(X_i)\Theta_i + c_j(X_i),\, \sigma_j^2(X_i)
\right).
\label{eq:agg_likelihood}
\end{equation}

\paragraph{Prior Term.}
The contextual prior over the latent variable is Gaussian:
\begin{equation}
p(\Theta_i \mid X_i)
=
\mathcal{N}\!\left(
\Theta_i;\, \mu_0(X_i),\, \sigma_0^2(X_i)
\right).
\label{eq:agg_prior}
\end{equation}

\paragraph{Posterior Form.}
By Bayes' rule, the posterior distribution is proportional to the product
of the likelihood in Eq.~\eqref{eq:agg_likelihood} and the prior in Eq.~\eqref{eq:agg_prior}:
\begin{equation}
p(\Theta_i \mid B_i, X_i)
\propto
p(B_i \mid \Theta_i, X_i)\, p(\Theta_i \mid X_i).
\end{equation}

Since both terms are Gaussian in $\Theta_i$, the posterior is also Gaussian.
Collecting terms in the exponent yields
\begin{equation}
p(\Theta_i \mid B_i, X_i)
=
\mathcal{N}\!\left(
\Theta_i;\,
\mu_{\mathrm{post}}(B_i,X_i),\,
\sigma_{\mathrm{post}}^2(B_i,X_i)
\right),
\end{equation}
where the posterior precision is given by
\begin{equation}
\sigma_{\mathrm{post}}^{-2}(B_i,X_i)
=
\sigma_0^{-2}(X_i)
+
\sum_{j=1}^L
\frac{a_j^2(X_i)}{\sigma_j^2(X_i)},
\label{eq:agg_precision}
\end{equation}
and the posterior mean is
\begin{equation}
\mu_{\mathrm{post}}(B_i,X_i)
=
\sigma_{\mathrm{post}}^2(B_i,X_i)
\left[
\frac{\mu_0(X_i)}{\sigma_0^2(X_i)}
+
\sum_{j=1}^L
\frac{a_j(X_i)\left(b_{ij}-c_j(X_i)\right)}{\sigma_j^2(X_i)}
\right].
\label{eq:agg_mean}
\end{equation}

\paragraph{Interpretation.}
This posterior provides a closed-form, interpretable combination of all sensor readings at site i: the posterior mean $\mu_{\mathrm{post}}$ corresponds to a reliability-weighted, bias-corrected fusion of all measurements. 
Each sensor contributes according to its calibration coefficient $a_j(X_i)$ and reliability $\sigma_j^2(X_i)^{-1}$: sensors with larger effective precision $a_j^2(X_i)/\sigma_j^2(X_i)$ exert greater influence on the posterior mean, while high-variance or weakly calibrated sensors are automatically down-weighted. The posterior variance $\sigma_{\mathrm{post}}^2$ quantifies epistemic uncertainty arising from limited sensor agreement and prior uncertainty.

\subsection{Posterior Predictive Derivation}

We now derive the predictive distribution for the true (unobserved) target variable $Y_i$
conditioned on the observed measurements $B_i$ and covariates $X_i$.

From the generative model, the conditional distribution of $Y_i$ given the latent variable
$\Theta_i$ is Gaussian:
\begin{equation}
p(Y_i \mid \Theta_i)
=
\mathcal{N}\!\left(Y_i;\, \Theta_i,\, \sigma_y^2\right),
\label{eq:pred_likelihood}
\end{equation}
where $\sigma_y^2$ captures irreducible aleatoric variability in the underlying process.

Given the posterior distribution of $\Theta_i$ derived in Section~\ref{sec:agg_posterior},
\begin{equation}
p(\Theta_i \mid B_i, X_i)
=
\mathcal{N}\!\left(
\Theta_i;\,
\mu_{\mathrm{post}}(B_i,X_i),\,
\sigma_{\mathrm{post}}^2(B_i,X_i)
\right),
\label{eq:theta_posterior}
\end{equation}
the posterior predictive distribution of $Y_i$ is obtained by marginalizing out $\Theta_i$:
\begin{equation}
p(Y_i \mid B_i, X_i)
=
\int p(Y_i \mid \Theta_i)\, p(\Theta_i \mid B_i, X_i)\, d\Theta_i.
\label{eq:pred_integral}
\end{equation}

Substituting Eqs.~\eqref{eq:pred_likelihood} and \eqref{eq:theta_posterior},
the integral in Eq.~\eqref{eq:pred_integral} corresponds to the convolution of two Gaussian
distributions and admits a closed-form solution:
\begin{equation}
p(Y_i \mid B_i, X_i)
=
\mathcal{N}\!\left(
Y_i;\,
\mu_{\mathrm{post}}(B_i,X_i),\,
\sigma_{\mathrm{post}}^2(B_i,X_i) + \sigma_y^2
\right).
\label{eq:posterior_predictive}
\end{equation}

\paragraph{Uncertainty Decomposition.}
The predictive variance decomposes additively into two components:
\begin{equation}
\mathrm{Var}(Y_i \mid B_i, X_i)
=
\underbrace{\sigma_{\mathrm{post}}^2(B_i,X_i)}_{\text{epistemic uncertainty}}
+
\underbrace{\sigma_y^2}_{\text{aleatoric uncertainty}}.
\end{equation}

The epistemic term reflects uncertainty in the latent variable $\Theta_i$
due to limited or conflicting sensor information,
while the aleatoric term captures irreducible variability in the true process.

Finally, the posterior predictive mean provides the fused point estimate:
\begin{equation}
\hat{Y}_i^{\text{fused}} = \mathbb{E}[Y_i \mid B_i, X_i] = \mu_{\text{post}}(B_i,X_i),
\label{eq:fused_mean}
\end{equation}
which is used for downstream analysis and calibration.
The predictive variance quantifies its total uncertainty:
\begin{equation}
\mathrm{Var}(Y_i \mid B_i, X_i)
= \sigma_{\text{post}}^2(B_i,X_i) + \sigma_y^2.
\label{eq:fused_var}
\end{equation}

These two quantities form the foundation of the subsequent Monte Carlo
Conformal Prediction (MC-CP) calibration procedure, which converts
the model-based predictive uncertainties into empirically validated
prediction intervals with guaranteed coverage.

\section{Additional Discussion on Identifiability}
\label{app:identifiability}

Affine non-identifiability is common in latent-variable models,
including factor analysis, item response theory, and multi-instrument calibration,
where scale and offset constraints are required for interpretability.

In the unconstrained model, the likelihood depends on $\Theta_i$
only through $a_j(X_i)\Theta_i + c_j(X_i)$.
Thus, scale and location transformations of $\Theta_i$
can be absorbed by the bias parameters without changing $p(B_i \mid X_i)$.
This ambiguity is structural and persists even with infinite data.

Consequently, likelihood-based training alone cannot distinguish between:
(i) a large-scale latent variable with small sensor gains, and
(ii) a small-scale latent variable with large sensor gains.
Additive shifts in $\Theta_i$ can likewise be absorbed into $c_j(X_i)$.

Empirically, we observe that this degeneracy manifests as decreasing validation NLL while predictive RMSE increases. Anchoring removes this degree of freedom and stabilizes optimization.

\paragraph{Effect of Sensor Anchoring}

\begin{figure}[t]
  \centering
  \includegraphics[width=\linewidth]{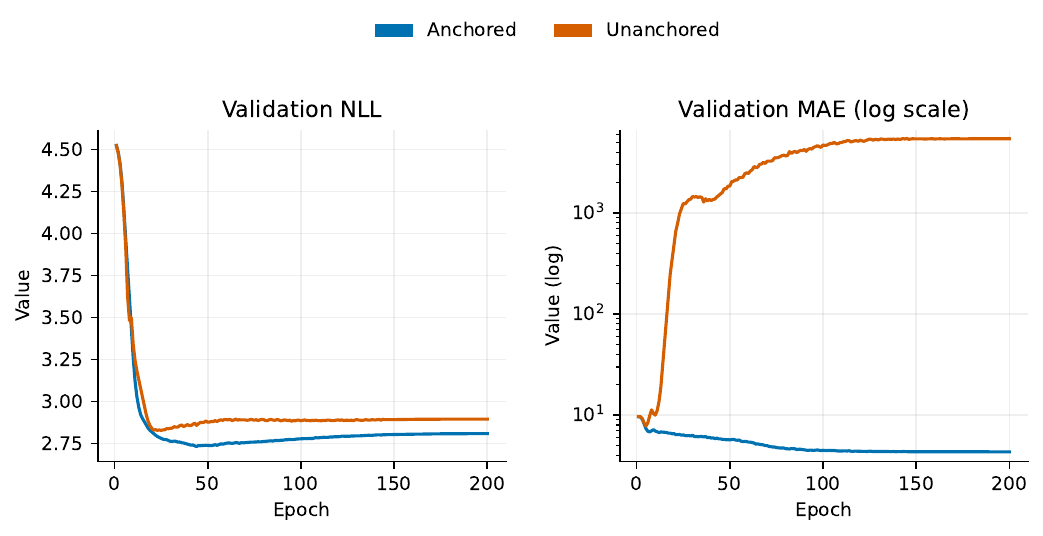}
  \caption{Validation NLL and MAE for unanchored and anchored models. Without anchoring, NLL decreases while MAE increases due to scale drift. Anchoring restores aligned optimization behavior.}
  \label{fig:anchored-unanchored}
\end{figure}

Figure~\ref{fig:anchored-unanchored} illustrates the above-mentioned optimization degeneracy and its resolution under anchoring: the unanchored model shows decreasing NLL concurrent with increasing MAE, whereas the anchored model exhibits aligned improvements in both metrics.

\section{Temporal Block Splitting Details}
\label{app:temporal_split}

\subsection{Block Construction}

Let $\{t_1, \dots, t_T\}$ denote ordered timestamps.
We partition the timeline into contiguous blocks of length $B$:
\[
\mathcal{I}_k = \{ t_{s_k}, \dots, t_{s_k + B - 1} \},
\]
with separation constraint
\[
s_{k+1} \ge s_k + B + G,
\]
where $G$ is the temporal gap (36 hours).
Entire blocks are assigned to splits, ensuring no temporal overlap.

\subsection{Weak Dependence Interpretation}

Strict exchangeability does not hold under temporal dependence.
If the data-generating process is $\alpha$-mixing with mixing coefficient $\alpha(\tau) \to 0$ as $\tau \to \infty$, then sufficiently separated observations become approximately independent.

By introducing a gap $G$, we reduce cross-block dependence and improve the plausibility of approximate exchangeability at the block level. Conformal coverage should therefore be interpreted as approximate under weak dependence.

\subsection{Practical Considerations}

Block splitting:
\begin{itemize}
    \item mitigates leakage from short-range autocorrelation,
    \item preserves seasonal variability across splits,
    \item yields empirically stable coverage.
\end{itemize}

\section{Residual Autocorrelation Analysis for Gap Selection}
\label{app:acf_gap_selection}

\begin{figure}[thb]
\centering
\includegraphics[width=\linewidth]{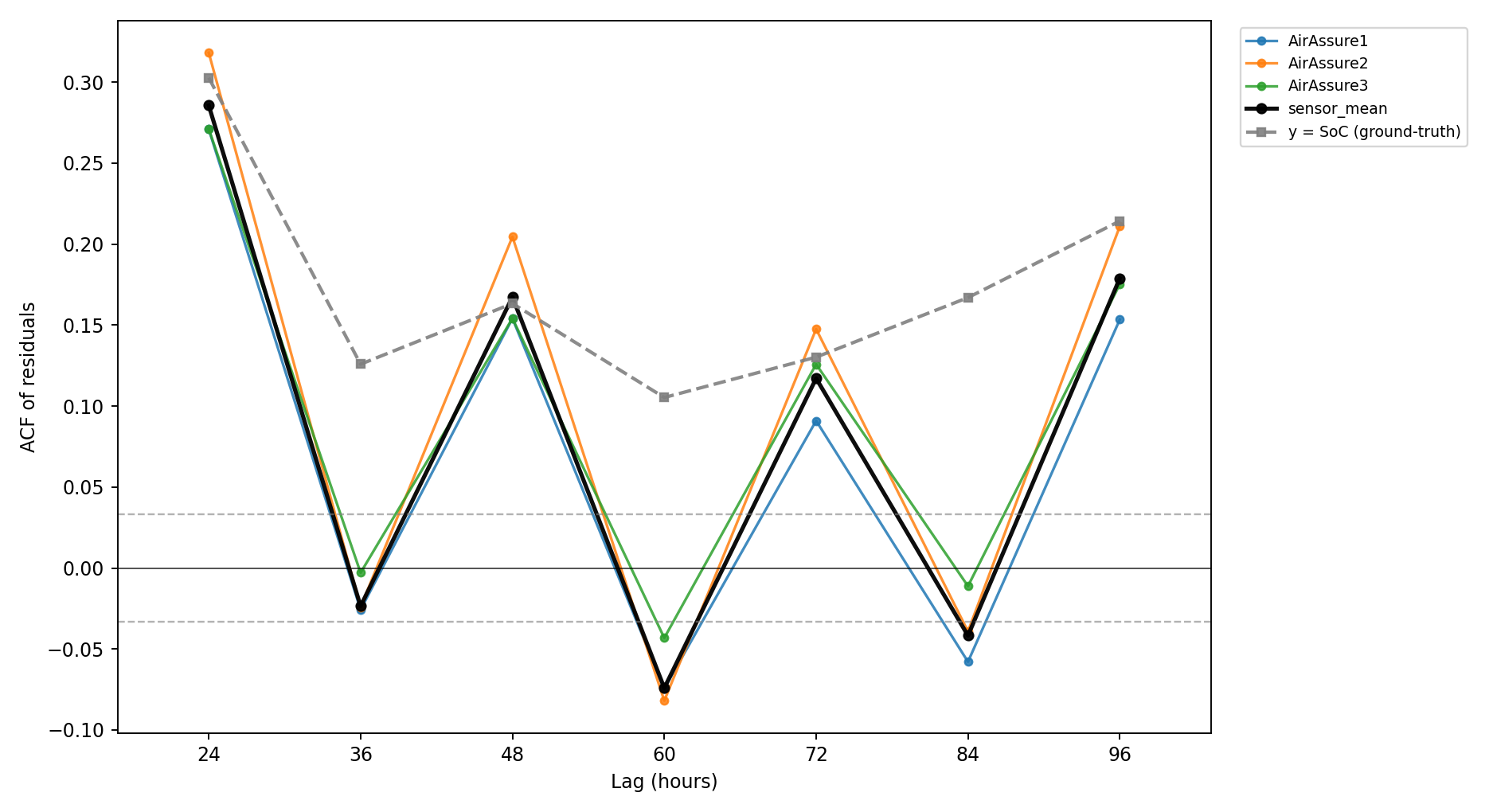}
\caption{Residual autocorrelation at selected lags for CAIRSENSE. Autocorrelation enters the 95\% confidence band at 36 hours, motivating the choice of a 36-hour gap between temporal blocks.}
\label{fig:acf_gap_selection}
\end{figure}

Figure~\ref{fig:acf_gap_selection} reports the autocorrelation function (ACF) of residuals at selected lags for the CAIRSENSE dataset. Residuals are computed on the training split as the difference between individual sensor measurements and the reference target column. At a 24-hour lag, residual autocorrelation remains outside the 95\% confidence band, reflecting strong diurnal structure. At 36 hours, the ACF falls within the confidence band, indicating that short-range temporal dependence has effectively decayed. Although autocorrelation re-emerges at longer lags (e.g., 48 and 72 hours), these correspond to multi-day periodic effects rather than short-range dependence that would induce leakage between adjacent temporal blocks. We therefore select a 36-hour gap as the smallest separation for which residual dependence becomes statistically indistinguishable from zero, balancing temporal independence with data efficiency.



\section{Toy Spatial Data Generation Process}
\label{app:toy-data-generation}

We generate a synthetic spatial regression dataset with a latent ground-truth field and three biased/noisy sensors. The construction is designed to induce heterogeneous sensor reliability, multiplicative/additive bias, heteroscedastic noise, and rare outliers.

\paragraph{Overview.}
For each sample \(i=1,\dots,N\), we first draw a spatial location, construct periodic covariates, generate a latent ground-truth target \(Y_i^\star\), and then generate three sensor observations \(\{B_{i0},B_{i1},B_{i2}\}\) from distinct corruption mechanisms.

Let \(\gamma>0\) denote an \emph{ambiguity factor} controlling the overall severity of sensor disagreement (bias/noise amplitude).

\paragraph{Step 1: Spatial locations and normalized coordinates.}
Let the spatial domain be
\[
\mathcal{D}=[\lambda_{\min},\lambda_{\max}] \times [\phi_{\min},\phi_{\max}].
\]
For each \(i\),
\begin{equation}
\lambda_i \sim \mathcal{U}(\lambda_{\min},\lambda_{\max}),
\qquad
\phi_i \sim \mathcal{U}(\phi_{\min},\phi_{\max}),
\end{equation}
independently.

Define center and isotropic scale
\begin{equation}
c_\lambda=\frac{\lambda_{\min}+\lambda_{\max}}{2},
\qquad
c_\phi=\frac{\phi_{\min}+\phi_{\max}}{2},
\qquad
s=\frac{1}{2}\max\!\big(\lambda_{\max}-\lambda_{\min},\,\phi_{\max}-\phi_{\min}\big),
\end{equation}
and normalized coordinates
\begin{equation}
\tilde{\lambda}_i=\frac{\lambda_i-c_\lambda}{s},
\qquad
\tilde{\phi}_i=\frac{\phi_i-c_\phi}{s}.
\end{equation}

\paragraph{Step 2: Covariates.}
We define periodic spatial covariates
\begin{equation}
X_i=
\begin{bmatrix}
\sin(2\pi \tilde{\lambda}_i) \\
\cos(2\pi \tilde{\lambda}_i) \\
\sin(2\pi \tilde{\phi}_i) \\
\cos(2\pi \tilde{\phi}_i)
\end{bmatrix}
\in \mathbb{R}^4.
\end{equation}

\paragraph{Step 3: Ground-truth latent field.}
Define a smooth background component
\begin{equation}
g_i
=
0.3\sin(2\pi\tilde{\lambda}_i)
+0.25\cos(2\pi\tilde{\phi}_i)
+0.1\sin(2\pi\tilde{\lambda}_i)\cos(2\pi\tilde{\phi}_i).
\end{equation}

To introduce localized spatial structure, sample \(K=2\) hotspot parameters (once per generated dataset):
\begin{align}
m_{\lambda,k}, m_{\phi,k} &\sim \mathcal{U}(-0.3,1.3), \\
A_k &\sim \mathcal{U}(2,5), \\
\tau_k &\sim \mathcal{U}(0.06,0.15),
\qquad k=1,2,
\end{align}
independently.

Then the ground-truth target is generated as
\begin{equation}
Y_i^\star
=
\Bigg[
15 + 5g_i
+ \sum_{k=1}^{2}
A_k
\exp\!\left(
-\frac{(\tilde{\lambda}_i-m_{\lambda,k})^2+(\tilde{\phi}_i-m_{\phi,k})^2}{\tau_k}
\right)
+ \epsilon_i
\Bigg]_+,
\label{eq:toy_true_field}
\end{equation}
where
\begin{equation}
\epsilon_i \sim \mathcal{N}(0,\sigma_{\mathrm{spatial}}^2),
\qquad
[z]_+ := \max(z,0).
\end{equation}

\paragraph{Step 4: Sensor observations.}
We generate three sensors \(B_{i0}, B_{i1}, B_{i2}\), each with a distinct corruption mechanism. All sensor outputs are truncated to be nonnegative via \([\,\cdot\,]_+\).

\subparagraph{Sensor 0 (highest reliability).}
Sample dataset-level calibration parameters
\begin{equation}
\alpha_0 \sim \mathcal{N}\!\big(1,(0.02\gamma)^2\big),
\qquad
\beta_0 \sim \mathcal{N}\!\big(0,(0.2\gamma)^2\big).
\end{equation}
For each \(i\),
\begin{align}
\eta_{i0} &\sim \mathcal{N}\!\big(0,(0.8\gamma)^2\big), \\
B_{i0}
&=
\Big[
\alpha_0 Y_i^\star + \beta_0
+ 0.15\gamma\,\sin(4\pi\tilde{\lambda}_i)\cos(4\pi\tilde{\phi}_i)
+ \eta_{i0}
\Big]_+.
\end{align}

\subparagraph{Sensor 1 (moderate noise, heteroscedastic).}
Sample a dataset-level additive bias
\begin{equation}
\beta_1 \sim \mathcal{N}\!\big(0,(0.5\gamma)^2\big).
\end{equation}
For each \(i\), define a heteroscedastic noise scale
\begin{equation}
\sigma_{i1}=(1+0.02Y_i^\star)\gamma,
\end{equation}
and sample
\begin{align}
\eta_{i1} &\sim \mathcal{N}(0,\sigma_{i1}^2), \\
B_{i1}
&=
\Big[
1.2\,Y_i^\star + \beta_1
+ 0.4\gamma\,\cos(6\pi\tilde{\phi}_i)
+ \eta_{i1}
\Big]_+.
\end{align}

\subparagraph{Sensor 2 (lowest reliability, with outliers).}
Sample a dataset-level additive bias
\begin{equation}
\beta_2 \sim \mathcal{N}\!\big(0,(0.7\gamma)^2\big).
\end{equation}
For each \(i\), define
\begin{equation}
\sigma_{i2}=(1.3+0.03Y_i^\star)\gamma,
\end{equation}
and generate noise with rare outliers:
\begin{align}
\varepsilon_{i2} &\sim \mathcal{N}(0,\sigma_{i2}^2), \\
o_i &\sim \mathrm{Bernoulli}(0.05), \\
\zeta_{i2} &\sim \mathcal{N}\!\big(0,(3\sigma_{i2})^2\big), \\
\eta_{i2} &= \varepsilon_{i2} + o_i\zeta_{i2}.
\end{align}
Then
\begin{equation}
B_{i2}
=
\Big[
1.4\,Y_i^\star + \beta_2
+ 0.5\gamma\,\sin(5\pi\tilde{\lambda}_i\tilde{\phi}_i)
+ \eta_{i2}
\Big]_+.
\end{equation}

\paragraph{Resulting dataset.}
The synthetic dataset consists of tuples
\[
\{(X_i, B_{i0}, B_{i1}, B_{i2}, Y_i^\star)\}_{i=1}^N,
\]
where \(X_i\) are observable covariates, \((B_{i0},B_{i1},B_{i2})\) are the three sensor readings, and \(Y_i^\star\) is used only for evaluation.

\newpage
\section{Summary of Key notation}\label{app:notation}

\begin{table}[b!]
  \centering
  \caption{Summary of key notation}\label{tab:notation}
  \small
  \begin{tabular}{@{}l p{0.74\linewidth}@{}}
    \toprule
    \bfseries Symbol & \bfseries Description \\
    \midrule
    \multicolumn{2}{@{}l}{\textit{Problem setup and NCAM}}\\
    $N$ & Number of entities (data points).\\
    $i, j$ & Entity and sensor indices.\\
    $L$ & Total number of sensors.\\
    $X_i$ & Contextual covariates for entity $i$.\\
    $B_i$ & Vector of sensor readings for entity $i$.\\
    $\Theta_i$ & Latent fused target variable.\\
    $Y_i^\star$ & Ground-truth target (evaluation only).\\
    $\mu_0, \sigma_0^2$ & Prior mean and variance.\\
    $a_j, c_j$ & Multiplicative and additive sensor bias.\\
    $\sigma_j^2$ & Sensor-specific noise variance (reliability).\\
    $\sigma_y^2$ & Irreducible aleatoric variance.\\
    $\mu_{\text{post}}, \sigma_{\text{post}}^2$ & Posterior mean and variance.\\
    $\hat{Y}_i^{\text{fused}}$ & Fused point prediction.\\
    $\hat{\sigma}_{i,\text{pred}}^2$ & Total predictive variance.\\
    $j^\star$ & Index of the anchor sensor.\\
    $\lambda_0$ & Variance regularization weight for prior variance penalty.\\
    $\lambda_s$ & Variance regularization weight for sensor variance penalty.\\
    $\mathcal{R}_{\mathrm{var}}$ & Variance regularization term added to the training objective.\\
    $\mathcal{L}_{\mathrm{total}}$ & Total training objective, $\mathcal{L}_{\mathrm{NCAM}} + \mathcal{R}_{\mathrm{var}}$.\\
    \midrule
    \multicolumn{2}{@{}l}{\textit{Conformal calibration}}\\
    $\alpha$ & Target miscoverage level.\\
    $m, r$ & Number of samples and sample index.\\
    $s_i^{(r)}$ & Normalized conformity score.\\
    $q_{1-\alpha}$ & Empirical quantile of conformity scores.\\
    $[L_i, U_i]$ & CP prediction interval.\\
    $\tilde{\Theta}_{ij}$ & Bias-corrected sensor estimate.\\
    $s_{ij}^{\mathrm{sens}}$ & Sensor-anchored conformity score.\\
    $I_i^{\mathrm{sens}}$ & Sensor-anchored prediction interval.\\
    \bottomrule
  \end{tabular}
\end{table}

\end{document}